\pdfoutput=1

\documentclass[11pt]{article}

\usepackage[preprint]{acl}

\usepackage{times}
\usepackage{latexsym}

\usepackage[T1]{fontenc}

\usepackage[utf8]{inputenc}

\usepackage{microtype}

\usepackage{inconsolata}

\usepackage[utf8]{inputenc} 
\usepackage[T1]{fontenc}    
\usepackage{hyperref}       
\usepackage{url}            
\usepackage{booktabs}       
\usepackage{multirow}
\usepackage{amsfonts}       
\usepackage{nicefrac}       
\usepackage{microtype}      
\usepackage{xcolor}         
\usepackage{graphicx}
\usepackage{wrapfig}
\usepackage{listings}
\usepackage{caption}
\usepackage{subcaption}
\usepackage{amsmath}

\definecolor{codegreen}{rgb}{0,0.6,0}
\definecolor{codegray}{rgb}{0.5,0.5,0.5}
\definecolor{codepurple}{rgb}{0.07,0,0.53}
\definecolor{codered}{RGB}{189,41,0}
\definecolor{codecomment}{RGB}{153,153,153}
\definecolor{backcolour}{rgb}{0.96,0.96,0.96}
\definecolor{mygreen}{rgb}{0.0, 0.5, 0.0}

\lstdefinestyle{mystyle}{
    backgroundcolor=\color{backcolour},   
    commentstyle=\color{codegreen},
    keywordstyle=\color{codered},
    numberstyle=\tiny\color{codegray},
    stringstyle=\color{codepurple},
    emph={neumann_iterations, steps, problems, dependencies},          
    emphstyle=\color{codered},    
    basicstyle=\ttfamily\tiny,
    breakatwhitespace=false,         
    breaklines=true,                 
    captionpos=b,                    
    keepspaces=true,                 
    numbersep=5pt,                  
    showspaces=false,                
    showstringspaces=false,
    showtabs=false,   
    morekeywords={>,<,.,;,-,!,=,~},
    tabsize=2
}
\newcommand*\inlinelargeimage[1]{\raisebox{-0.15\baselineskip}{$\,$\includegraphics[height=1.3\baselineskip]{#1}$\,\,$}}
\newcommand{\redcoicon}{\inlinelargeimage{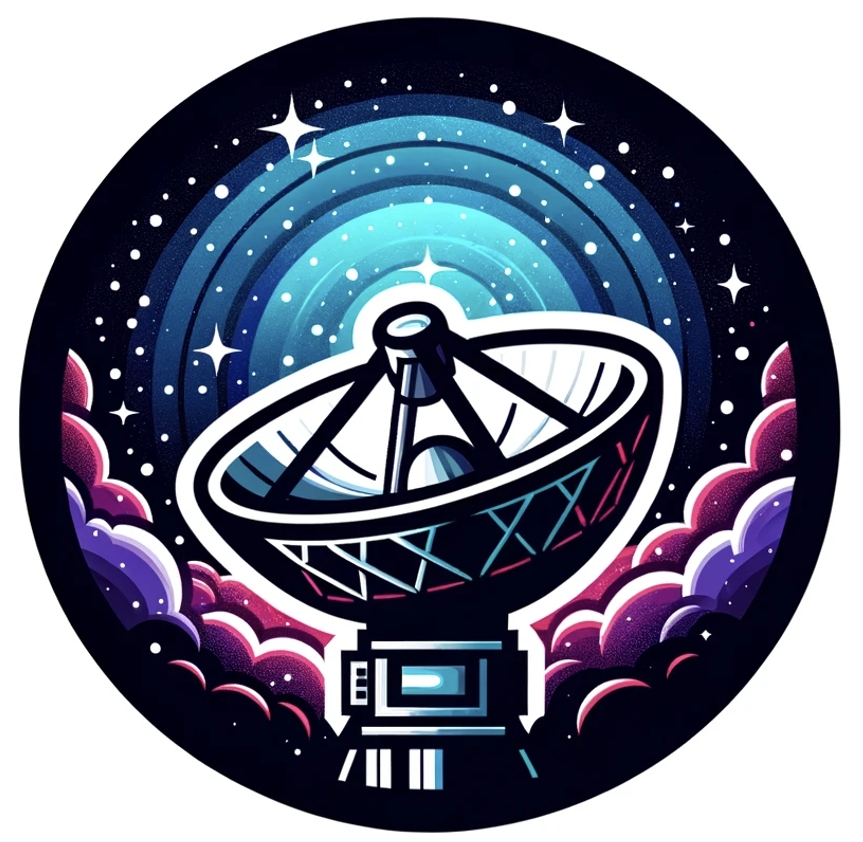}}

\title{\redcoicon RedCoast: A Lightweight Tool to Automate Distributed Training \\ of LLMs on Any GPU/TPUs}

\author{%
  Bowen Tan$^{1}$, Yun Zhu$^{2}$, Lijuan Liu$^{2}$, Hongyi Wang$^{1}$, Yonghao Zhuang$^{1}$, \\
  {\bf Jindong Chen$^{2}$, Eric Xing$^{1, 3, 5}$, Zhiting Hu$^{4}$} \\
  $^1$Carnegie Mellon University,~~ $^2$Google Research,~~ $^3$Petuum Inc.,~~ $^4$UC San Diego, \\
  $^5$Mohamed bin Zayed University of Artificial Intelligence \\
}

\begin{document}

\maketitle

\begin{abstract}
The recent progress of AI can be largely attributed to large language models (LLMs). However, their escalating memory requirements introduce challenges for machine learning (ML) researchers and engineers. Addressing this requires developers to partition a large model to distribute it across multiple GPUs or TPUs. This necessitates considerable coding and intricate configuration efforts with existing model parallel tools, such as Megatron-LM, DeepSpeed, and Alpa. These tools require users' expertise in machine learning systems (MLSys), creating a bottleneck in LLM development, particularly for developers without MLSys background. In this work, we present \textit{RedCoast (Redco)}, a lightweight and user-friendly tool crafted to automate distributed training and inference for LLMs, as well as to simplify ML pipeline development. The design of Redco emphasizes two key aspects. Firstly, to automate model parallelism, our study identifies two straightforward rules to generate tensor parallel strategies for any given LLM. Integrating these rules into Redco facilitates effortless distributed LLM training and inference, eliminating the need of additional coding or complex configurations. We demonstrate the effectiveness by applying Redco on a set of LLM architectures, such as GPT-J, LLaMA, T5, and OPT, up to the size of 66B. Secondly, we propose a mechanism that allows for the customization of diverse ML pipelines through the definition of merely three functions, avoiding redundant and formulaic code like multi-host related processing. This mechanism proves adaptable across a spectrum of ML algorithms, from foundational language modeling to complex algorithms like meta-learning and reinforcement learning. As a result, Redco implementations exhibit significantly fewer lines of code compared to their official counterparts. \footnote{\texttt{RedCoast (Redco)} has been released under Apache 2.0 license at \url{https://github.com/tanyuqian/redco}.}
\end{abstract}

\vspace{-15pt}
\section{Introduction}
\vspace{-5pt}

In recent years, the field of AI has witnessed profound advancements, predominantly attributed to the advent of LLMs with an impressive number of parameters, spanning from billions to hundreds of billions \cite{zhao2023survey}. Notable examples include GPT-4 \cite{OpenAI2023GPT4TR} and LLaMA \cite{Touvron2023LLaMAOA}. Yet, the size of these LLMs presents distinct challenges in terms of model deployment for ML researchers and engineers. The primary challenge arises from the substantial memory requirements of LLMs, often exceed the capability of a single GPU or TPU. This necessitates the use of model parallelism, a technique that partitions the LLMs into various shards, subsequently distributing them across multiple devices or even different hosts. However, achieving this partitioning requires intricate engineering, including the formulation of a tensor-specific splitting strategy. While several specialized tools like DeepSpeed \cite{rasley2020deepspeed}, Alpa \cite{zheng2022alpa}, and FSDP \cite{zhao2023pytorch} provide diverse model parallelism solutions, but they demand significant additional coding and intricate configurations based on model architecture and hardware specifics, requiring in-depth understanding of MLSys. Such additional efforts make the deployment of LLMs particularly challenging, especially for users without MLSys expertise, such as algorithm developers or researchers. At times, the intricacy of coding for model parallelism proves to be even more daunting than the algorithm design itself.

\begin{figure*}
    \centering
    \includegraphics[width=0.95\textwidth]{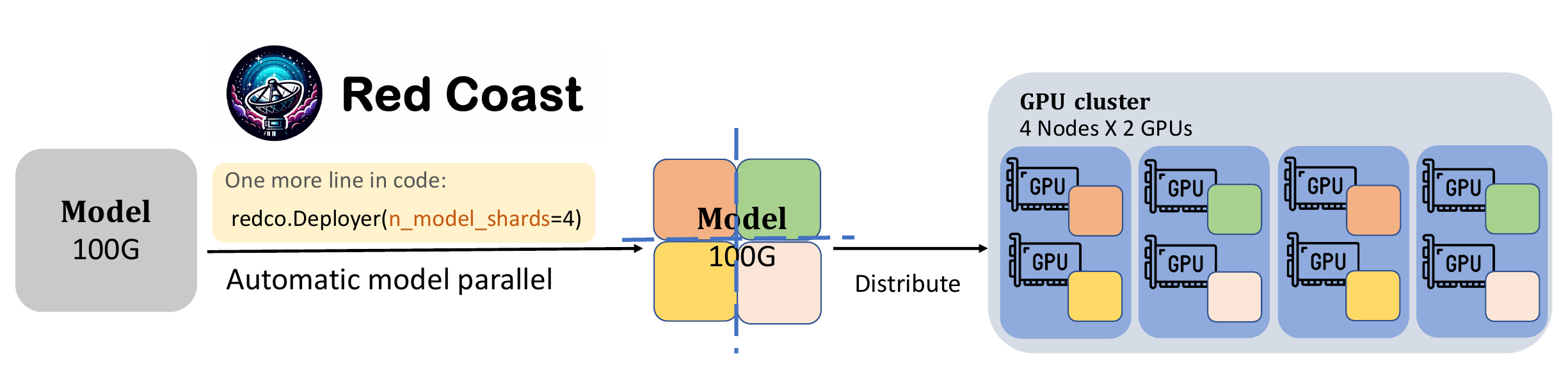}
    \vspace{-10pt}
    \caption{With a number of shards specified by user, Redco automatically conduct the model partitioning and distribution across hosts and devices.}
    \label{fig:my_label}
    \vspace{-15pt}
\end{figure*}

In this work, we introduce \textit{RedCoast (Redco)}\footnote{For simplicity, we will use Redco more frequently in the rest of this paper.}, a lightweight and user-friendly tool designed to automate the distributed training and inference of LLMs, thereby users without MLSys expertise can also effortlessly use the tool without additional coding or intricate configurations. Furthermore, we propose a novel and neat mechanism to implement ML algorithms. This method necessitates users to define merely three functions as their pipeline design, with Redco managing all the remaining details in execution, such as data parallelism, multi-host related processing, checkpointing, etc.

Redco's design emphasizes two key aspects. The first is the automatic model parallelism. We identify two straightforward rules to generate the model parallel strategy for arbitrarily given transformer architecture, and integrate them into Redco. Unlike tools such as Megatron \cite{shoeybi2019megatron} and DeepSpeed \cite{rasley2020deepspeed} which require users to manipulate model forward function for different architecture and system specifics, Redco automates the process, where users only need to specify the desired number of shards to partition the model. We verified the effectiveness of Redco's model parallel strategy on multiple LLMs including LLaMA-7B \cite{Touvron2023LLaMAOA}, T5-11B \cite{raffel2020exploring}, and OPT-66B \cite{zhang2022opt}. Moreover, pipelines driven by Redco demonstrate efficiency superior to those implemented with FSDP \cite{zhao2023pytorch}, and closely matching the performance of Alpa \cite{zheng2022alpa}, the tool with state-of-the-art model parallel efficiency.

Another pivotal feature of Redco is the neat mechanism for ML pipeline development. With Redco, users only need to write three intuitive functions to define a ML pipeline: a collate function to convert raw data examples into model inputs (e.g., text tokenization); a loss function to execute the model and compute loss (e.g., cross-entropy); and a predict function to run the model and deliver outcomes (e.g., beam search). With the defined pipeline from a user, Redco automates all the remaining of pipeline execution such as data parallelism, multi-host related processing, checkpointing, log maintenance, and so forth. We demonstrate this neat mechanism is applicable to various ML paradigms, spanning from basic language modeling and sequence-to-sequence (seq2seq), to more complex algorithms like meta-learning and reinforcement learning. Redco-based implementations consistently exhibit substantially fewer lines of code compared to their official counterparts.

\vspace{-5pt}
\section{Related work}
\vspace{-5pt}

\paragraph{Distributed Machine Learning.}

Distributed machine learning refers to the utilization of multiple computing devices, typically GPUs or TPUs, for the efficient training and inference of ML models with large datasets or large models. It usually includes data parallelism and model parallelism. Data parallelism involves dividing a large dataset into multiple subsets, with each subset processed independently by a separate computing device, and every device maintains a full copy of the model parameters. However, data parallelism is limited in its ability to handle large models that exceed the memory constraints of individual devices. Model parallelism addresses this limitation by splitting and distributing the model across multiple devices, with each responsible for a portion of the model. Although it offers a solution for large models, model parallelism is more complex to implement than data parallelism due to the necessity of careful model partitioning. Tools such as Megatron-LM \cite{shoeybi2019megatron, narayanan2021efficient}, DeepSpeed \cite{rasley2020deepspeed}, FSDP \cite{zhao2023pytorch}, and Alpa \cite{zheng2022alpa}, have been developed to facilitate model parallelism. These tools support the model partitioning but still require significant coding and configuration efforts based on specific model architecture and hardware settings. In this work, Redco offers automatic data parallelism by default, and provides automatic model parallelism for LLMs, which is the majority of model parallelism use cases. Prioritizing user-friendliness, Redco enables users to execute distributed LLM training and inference by simply specifying the number of model shards for partitioning, without requiring users' MLSys expertise.

\paragraph{Pipeline development tools.} 

In the development process using neural network libraries such as PyTorch \cite{paszke2019pytorch} and Flax \cite{flax2020github}, certain boilerplate code is consistently present. Common operations, such as back-propagation, gradient application, and batch iteration, recur in nearly every ML pipeline. A variety of tools aim to streamline pipeline development by eliminating repetitive code while maintaining as much development flexibility as possible. PyTorchLightning \cite{lightning2019github} offers a default training loop within PyTorch, allowing users to customize their pipelines by inheriting a Trainer class and modifying hook functions such as loss function and checkpoint saving. However, this mechanism may not be intuitive for all users. For some people, it requires a learning curve to become comfortable. Furthermore, it may be unclear how to implement these hook functions for more complex algorithms, such as federated learning. HuggingFace-Transformers \cite{wolf2020transformers} provides a Trainer for PyTorch models, but it heavily relies on models defined in its specific transformer classes and primarily focuses on natural language processing pipelines. Keras \cite{chollet2015keras} delivers higher-level APIs on top of TensorFlow \cite{tensorflow2015-whitepaper}, enabling users to specify data, model, and loss functions. However, it is not well-suited for handling complex pipelines. Our proposed Redco is based on Flax, and uses a more intuitive and flexible approach for users to design their pipelines. This mechanism can be applied to a wide array of ML algorithms together with the automatic support for distributed training, including complex algorithms such as federated learning, meta-learning, and reinforcement learning.

\begin{table*}[t]
\small
    \centering
\vspace{-5pt}
\begin{tabular}{@{}r|l|l|l|l@{}}
\toprule
Server                  & 2 $\times$ 1080Ti        & 4 $\times$ A100        & 2 $\times$ TPU-v4                  & 16 $\times$ TPU-v4   \\ 
Device Memory          & 2 $\times$ 10G           & 4 $\times$ 40G         & 2 (hosts) $\times$ 4 (chips) $\times$ 32G & 16 $\times$ 4 $\times$ 32G  \\ \midrule
\multirow{2}{*}{Models} & BART-Large (1024) & LLaMA-7B (1024) & T5-XXL-11B (512)            & OPT-66B (512) \\
                        & GPT2-Large (512)  & GPT-J-6B (1024) & OPT-13B (1024)              &               \\ \bottomrule
\end{tabular}
\vspace{-5pt}
    \caption{Runnable model finetuning on different servers. Numbers inside the brackets are the maxinum length in training. All the settings are with full precision (fp32) with AdamW optimizer.}
    \vspace{-10pt}
    \label{tab:runnable}
\end{table*}

\section{Automatic Model Parallelism for LLMs}

\textit{Model parallelism} refers to distributing the computation of a large model across multiple GPUs or TPUs, in order to address the memory limitations of a single device. Two sub-paradigms within model parallelism are \textit{pipeline parallelism} and \textit{tensor parallelism}. Pipeline parallelism partitions the layers of the model across different devices, and tensor parallelism distributes every tensor in the model across multiple devices. 

Model parallel tools like Megatron or Alpa require a bunch of intricate configurations and extensively modifying users' code based on the model architecture and the hardware setting. For example, Megatron requires users rewriting the model forward function to customize the tensor sharding for tensor parallelism and annotate breakpoints for pipeline parallelism. This demands substantial MLSys expertise, which is not possessed by most algorithm developers or researchers. 

In this work, we develop an automatic model parallel strategy in Redco that applies across LLMs without requiring users' MLSys expertise or extra coding efforts.

\subsection{Rules to Automate Tensor Parallelism}

In Redco, we automate model parallelism via tensor parallelism. A tensor sharding strategy requires a dimension specified for each tensor. Along the dimension, the tensor is sharded and distributed across multiple devices. The objective of sharding strategy design is to minimize memory and runtime overhead associated with inter-device communication, which is usually brought by \texttt{reduce} or \texttt{gather} operations. For example, consider a tensor \(t\) is defined either as \(t = t_1 + t_2\) or \(t = (t_1, t_2)\) (concatenation), with \(t_1\) and \(t_2\) being stored on distinct devices (GPUs or TPUs). In this case, the computation of \(t\) requires message passing between the two devices.


Consider a dense layer in a neural network

\vspace{-10pt}
$$y = \sigma(xA)$$
\vspace{-15pt}

\noindent where $x$ denotes the input tensor, $\sigma$ is an element-wise activation function (e.g., ReLU, SiLU), and $A$ is the weight matrix, serving as the model parameter of the dense layer. \footnote{The bias term is omitted here because its computation is non-significant.} When the weight matrix $A$ is divided along its first dimension (dimension 0), a \texttt{reduce} operation becomes necessary to compute $y$. Formally,
$$
A = \begin{pmatrix}
{\color{cyan} A_1}\\
{\color{orange} A_2}
\end{pmatrix}
    \Longrightarrow
y = \sigma({\color{cyan} x_1 A_1} + {\color{orange} x_2 A_2})
$$
where $x_1$ and $x_2$ represent the first and second halves of the input tensor $x$'s dimensions, respectively. The computation on two devices are indicated by the colors. 

Conversely, when $A$ is partitioned along its second dimension (dimension 1), a \texttt{gather} operation is required to obtain $y$:
$$
A = ({\color{cyan} A_1}, {\color{orange} A_2})
    \Longrightarrow
y = (\sigma({\color{cyan} xA_1}), \sigma({\color{orange} xA_2}))
$$


Therefore, when the weight parameter of each dense layer is partitioned across an arbitrary dimension, reduce or gather operations might occur within every dense layer. However, by examining a pair of consecutive dense layers

\vspace{-10pt}
$$y = \sigma(\sigma(xA)B)$$
\vspace{-15pt}

\noindent where \(A\) and \(B\) denote the respective weights. By partitioning \(A\) and \(B\) across dimensions 1 and 0, it becomes feasible to execute these two layers with a single time of reduce operation:

\vspace{-20pt}
$$
\begin{aligned}
& A = ({\color{cyan} A_1}, {\color{orange} A_2}), \quad B = \begin{pmatrix}
{\color{cyan} B_1} \\
{\color{orange} B_2}
\end{pmatrix} \\
\Longrightarrow & y = \sigma(\sigma(xA)B) \\
& ~ = \sigma\left( \left(\sigma({\color{cyan} xA_1}), \sigma({\color{orange} xA_2})\right) \begin{pmatrix}
{\color{cyan} B_1} \\
{\color{orange} B_2}
\end{pmatrix}\right) \\
& ~ = \sigma( \sigma({\color{cyan} xA_1}) {\color{cyan} B_1} + \sigma({\color{orange} xA_2}) {\color{orange} B_2})
\end{aligned}
$$
\vspace{-10pt}

In this case, the operations \(\sigma({\color{cyan} xA_1}) {\color{cyan} B_1}\) and \(\sigma({\color{orange} xA_2}) {\color{orange} B_2}\) can be performed independently on separate devices, so that they only take a single reduce operation in the computation. 

Based on the observation above, we get heuristic insights in terms of inter-device communication on the transformer architecture, specifically within feed-forward and attention layers. For feed-forward layers, given the nature of matrix multiplication, dividing two consecutive matrices along different dimensions is expected to require less inter-device communication than if they are divided along a same dimension. For the attention layers, the output matrix $O$ is multiplied with each of the $Q, K, V$ matrices, so the matrix $O$ should be splitted along a dimension distinct from that chosen for $Q, K, V$. Based on these insights, we write two rules to determine the dimension along which to split each tensor in a model: 

\vspace{-5pt}
\begin{enumerate}
\item For fully-connected layers, alternate between splitting the parameter along dimension 1 and dimension 0.
\vspace{-5pt}
\item For attention layers, split $Q, K, V$ along dimension 0, and split the output projection matrix $O$ along dimension 1.
\end{enumerate}
\vspace{-5pt}

Leveraging the rules above enables Redco to devise a model parallel strategy tailored for any given LLM architecture. This enables the distributed training of LLMs with almost zero user effort. Users only specify the number of shards to split the given model, without additional coding or configuration efforts.

Note that our proposed rules are similar to a part of suggested configurations of Megatron \cite{shoeybi2019megatron}, but they don't summarize their separate configurations into rules, so that only a few LLM architectures (BERT, GPT, and T5) are supported in their implmentation\footnote{https://github.com/NVIDIA/Megatron-LM/tree/main}. To customize any new architectures under Megatron, users still have to rewrite the model's forward function and manually implement their model parallel strategy.


\vspace{-5pt}
\subsection{Implementation inside Redco} 

We implement tensor parallelism with the proposed strategy on top of \texttt{jax.pjit} function. This function compiles the computational graph and it merges operations on the same device to reduce unnecessary communication overhead \footnote{\url{https://jax.readthedocs.io/en/latest/notebooks/Distributed_arrays_and_automatic_parallelization.html}}. 

To integrate the proposed tensor parallel strategy, Redco has a function that takes in an arbitrary transformer architecture and produces a parameter sharding strategy based on the proposed rules. Moreover, in addition to automatically generating sharding strategies, Redco also enables their customization. This allows users with more advanced strategies to execute their approaches. Practical examples of the sharding strategies produced by Redco, applied to GPT-J and LLaMA, are showcased in the Appendix.


\vspace{-5pt}
\subsection{Evaluation}

\begin{figure*}
    \centering
    \includegraphics[width=\textwidth]{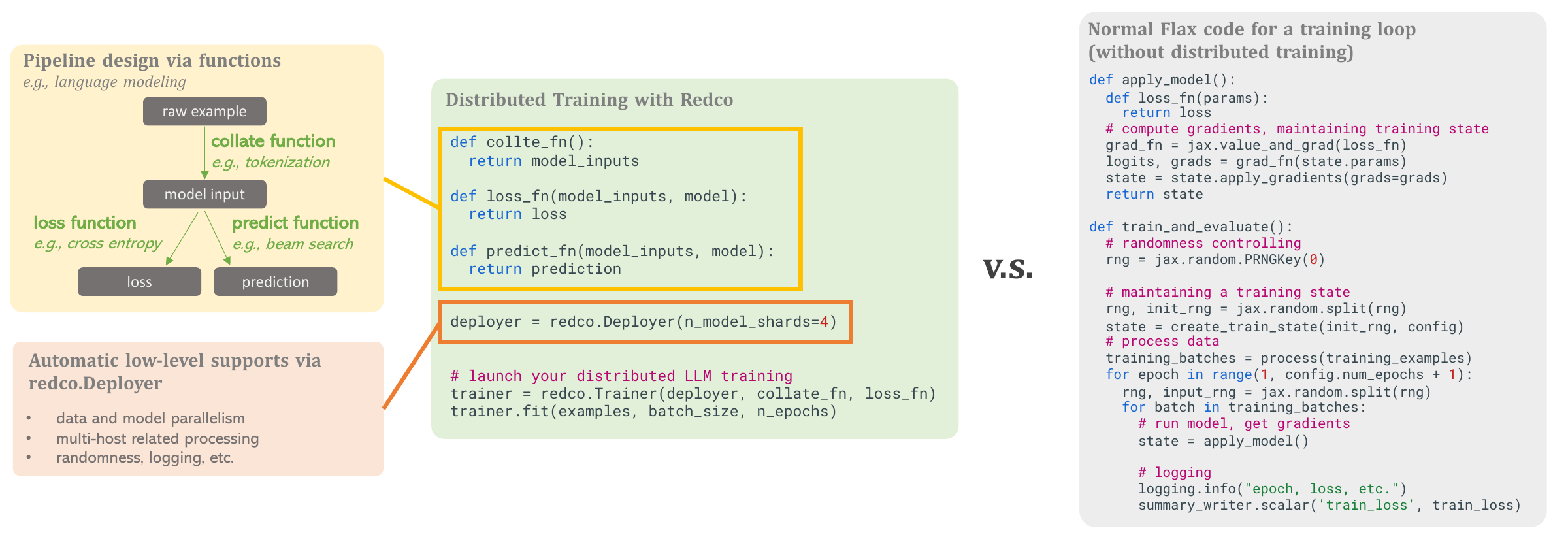}
    \vspace{-15pt}
    \caption{The template code of using Redco to implement distributed training, where users only have to design a pipepine through three fucntions, without concerning data and model parallelism, multi-host related processing, randomness control, etc., which eliminates a lot of boilerplate coding.}
    \label{fig:pipeline_design}
    \vspace{-10pt}
\end{figure*}

\begin{figure}[t]
    \centering
    \vspace{-5pt}
    \includegraphics[width=0.43\textwidth]{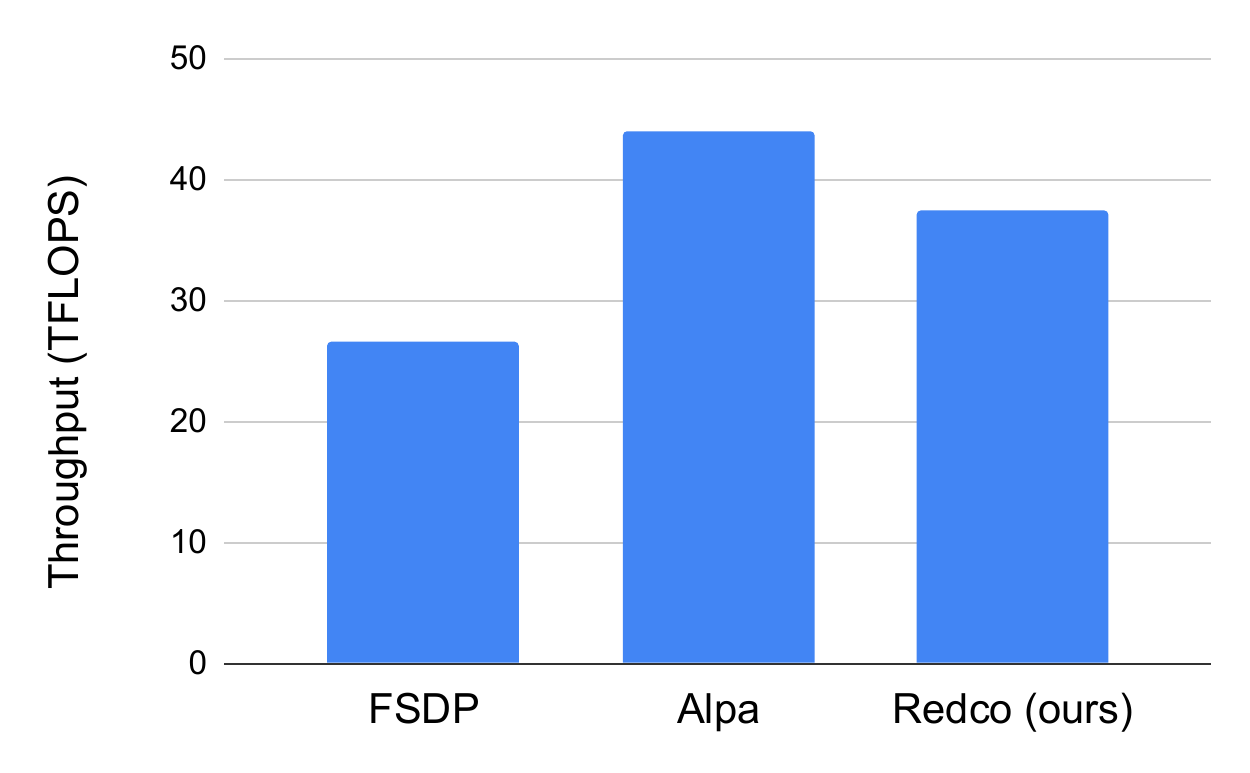}
    \vspace{-10pt}
    \captionof{figure}{The comparison of throughput in the running of GPT-J-6B on a $4 \times$ A100 server. Redco's performance surpasses that of FSDP and is close to that of Alpa, the tool with state-of-the-art model parallel efficiency but intricate engineering.}
    \label{fig:flops}
    \vspace{-15pt}
\end{figure}

\paragraph{Applicability test} We assess the applicability of our proposed automatic model parallel approach by applying Redco on an assorted collection of LLMs across various GPU and TPU servers. We conduct distributed training for LLMs without compromising the optimizer settings or precision. More precisely, we execute the distributed training under full precision (fp32) with the widely-used, yet memory-intensive, AdamW optimizer. We report operable LLMs on GPU and TPU servers with varying memory capacities.

The findings, as displayed in Table \ref{tab:runnable}, indicate that our straightforward, yet effective, automatic model parallel strategy is highly applcable across LLMs. For example, on small servers, such as those equipped 2 $\times$ 1080Ti, our strategy successfully runs large versions of BART and GPT-2 with text lengths up to 512 and 1024, respectively. On the larger servers such as the one with 16 TPU-v4 hosts, Redco effectively handles the training of the giant OPT-66B.

\vspace{-5pt}
\paragraph{Efficiency test}
We evaluate the efficiency of our proposed automatic model parallelism strategy in Redco on a server equipped 4 $\times$ A100 GPUs. We perform experiments by finetuning OPT-2.7B and GPT-J-6B, on the WikiText dataset, with full precision and AdamW optimizer. We compare Redco with two advanced model parallel tools: FSDP and Alpa.
The experimental results are summarized in Figure \ref{fig:flops}. The observed throughput reveals that Redco surpasses FSDP and is close to Alpa, the state-of-the-art model parallel tool. Notably, Alpa's implementation requires advanced MLSys expertise and significant coding efforts.

\begin{figure*}[t]
    \centering
\begin{lstlisting}[language=Python]
def collate_fn(raw_examples):
  return {
      "pixel_values": # pixel values of the images
      "token_ids": # token indicies of captions 
  }
  
def loss_fn(batch, params):
  logits = model(params, batch['pixel_values'], batch['token_ids']) # run model and get logits
  return cross_entropy(logits, batch['token_ids'])

def pred_fn(batch, params):
  return model.beam_search(params, batch['pixel_values'])
\end{lstlisting}
\vspace{-10pt}
    \caption{Pipeline design functions of image captioning under Redco.}
    \label{fig:img_cap_example}
\begin{lstlisting}[language=Python]
def collate_fn(raw_examples):
  return {
      "train_data": # a batch of few-shot training tasks
      "eval_data": # a batch of few-shot evaluation tasks
  }
  
def loss_fn(batch, params):
  params_inner = params - alpha * jax.grad(inner_loss)(params, batch['train_data'])
  return inner_loss(params_inner, batch['eval_data'])

def pred_fn(batch, params, model):
  params_inner = params - alpha * jax.grad(inner_loss)(params, batch['train_data'])
  return model(params_inner, batch['eval_data'])
\end{lstlisting}
\vspace{-10pt}
    \caption{Pipeline design functions of meta-learning (MAML) for few-shot learning under Redco. MAML's loss $L = \mathcal{L}(\mathcal{T}_{eval}, \theta')$ and $\theta' = \theta - \alpha \nabla_{\theta} \mathcal{L}(\mathcal{T}_{train}, \theta)$, where $\mathcal{T}_{train}$, $\mathcal{T}_{eval}$ refer to the data of training and evaluation tasks, and $\mathcal{L}(\cdot, \cdot)$ refers to the original loss function (\texttt{inner\_loss}), such as the cross-entropy loss for classification.}
    \label{fig:maml_example}
    \vspace{-10pt}
\end{figure*}

\section{RedCoast: Library Design}

In addition to the complexities of implementing model parallelism, ML pipelines often contain repetitive boilerplate code that demands significant effort from developers. Examples of such code include back-propagation, gradient application, batch iteration, and so forth. Furthermore, the hardware upgrades usually require patches on existing codebase. For example, a code developed within a single-GPU setting needs data parallelism and multi-host related processing to be added when adapted to multi-GPU machines or clusters.  

In Redco, we design a neat and user-friendly mechanism to simplify ML pipeline developments. Users only have to define their pipeline through three design functions, and Redco handles all the remaining of the pipeline execution. In this section, we will introduce the software design of Redco that enables this mechanism.

\subsection{Pipeline Design Through Three Functions}
\label{sec:three_funcs}
As shown in the yellow brick in Figure \ref{fig:pipeline_design}, in our proposed mechanism, every ML pipeline can be decoupled into three simple functions: \textit{collate function} to convert raw examples to model inputs, e.g., converting text sentences to be a batch of token indices via tokenization; \textit{loss function} to convert the model inputs to a scalar loss; and \textit{predict function} to convert the model inputs to the desired model outputs, such as running beam search for the language model. We demonstrate this framework with the implementation of image captioning and a meta-learning algorithm, MAML, as shown in Figure \ref{fig:img_cap_example} and Figure \ref{fig:maml_example}. These examples showcase that both simple and complex algorithms can be naturally defined under the proposed mechanism.

\subsection{Pipeline Execution with Automatic Low-level Supports}

For user-friendliness, there are merely three classes in Redco, i.e., \texttt{Deployer}, \texttt{Trainer}, and \texttt{Predictor}.
As shown in the orange brick in Figure \ref{fig:pipeline_design}, Redco streamlines the management of low-level and boilerplate processing in pipeline development through the \texttt{Deployer} class. This includes automatic model parallelism, as discussed in the previous section, as well as automatic data parallelism, multi-host related processing, checkpointing, and other convenient features such as randomness control and logging management.
The final execution of the pipeline is carried out by \texttt{Trainer} and \texttt{Predictor} of Redco. Supported by \texttt{Deployer}, they execute the training and inference of the pipeline defined by users via the functions as mentioned in Section \ref{sec:three_funcs}.
\footnote{We include a complete example in the Appendix implementing a distributed seq2seq pipeline with Redco.}

\begin{figure*}[t]
    \centering
    \vspace{-15pt}
    \includegraphics[width=0.9\textwidth]{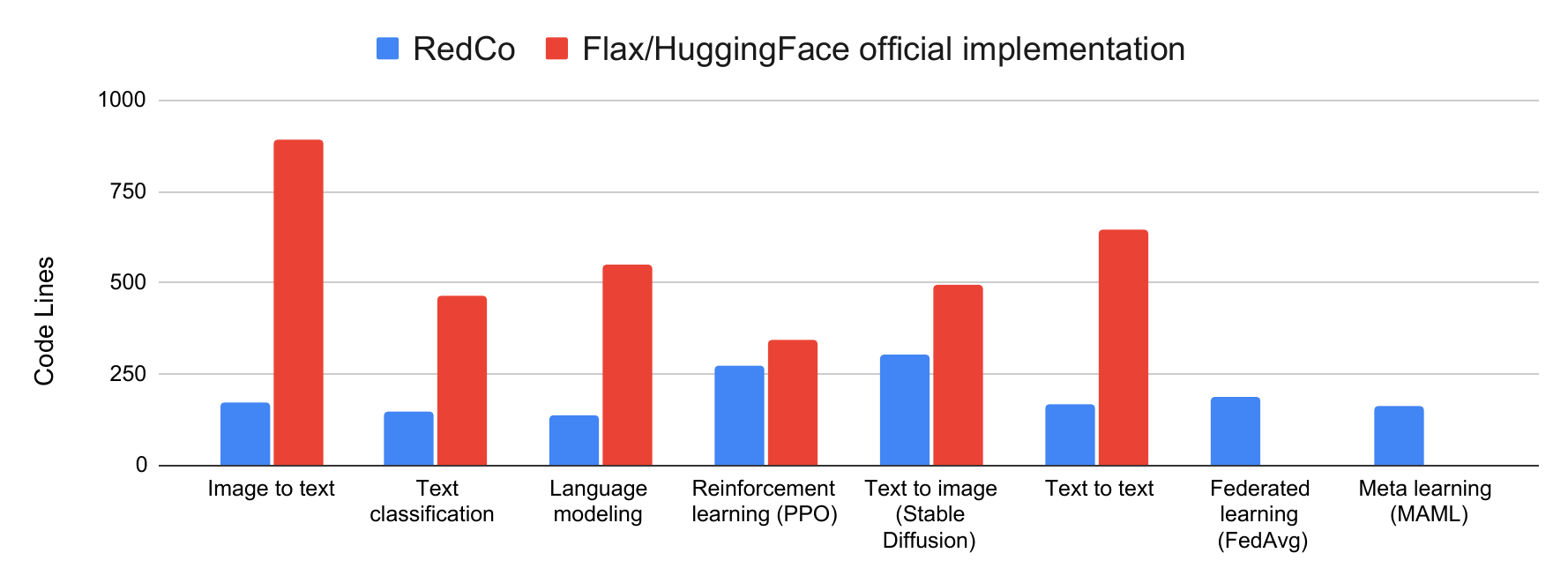}
    \vspace{-15pt}
    \caption{The comparison of code lines across a diverse set of ML algorithms. (There is no well-accepted official Flax implementations for FedAvg and MAML.)}
    \vspace{-15pt}
    \label{fig:lines}
\end{figure*}

\subsubsection{Multi-host Supports}
Large-scale distributed training typically involves intricate processes to accommodate multiple hosts. These processes include allocating data samples to each node and aggregating gradients or parameters across hosts, etc. Redco offers automatic support for multi-host environments and demonstrates compatibility with various platforms, including SLURM \cite{yoo2003slurm}, XManager\footnote{\url{https://github.com/google-deepmind/xmanager}}, as well as bare nodes interconnected via IP addresses. Notably, Redco allows users to maintain their existing pipeline design and execution code without additional efforts for multi-host environments.

\subsubsection{Checkpointing}
In distributed training frameworks, each typically employs a distinct formatting for checkpoint saving and loading, leading to a closed-loop system. For instance, Megatron utilizes a unique approach where model parameters and optimizer states are segmented based on the configuration of model parallelism. These checkpoints are inherently tied to Megatron, necessitating considerable effort for conversion into standard PyTorch checkpoint formats. Conversely, in Redco, we adopt a standardized and well-accepted checkpointing method. Here, both model parameters and optimizer states are encapsulated comprehensively within dictionaries of tensors. This approach is independent of the specific distributed training configurations, offering the advantage of simplicity in loading and utilization, even without Redco installation.

\subsubsection{Lightweight and Flexible Dependency}
Distributed training frameworks, such as Megatron and Alpa, often include modifications to foundational Python packages or CUDA kernels, resulting in stringent environment installation requirements. For instance, Alpa modifies the fundamental \texttt{jaxlib}\footnote{\url{https://pypi.org/project/jaxlib/}} library, thereby limiting its compatibility to jaxlib version 0.3.22 and CUDA version 11.3. They have been outdated compared to advanced versions of jaxlib 0.4.32 and CUDA 12.2, which are prevalent in many cluster environments today. In contrast, Redco is developed on top of Jax and Flax, without any modification to existing packages. Consequently, Redco is able to support a wider range of versions of Jax, Flax, and CUDA, in addition to accommodating various device types, including GPUs and TPUs.

\vspace{-5pt}
\subsection{Evaluation}
We implement a variety of machine learning paradigms using Redco, ranging from fundamental supervised learning techniques such as classification and regression, to more sophisticated algorithms including reinforcement learning and meta-learning. Figure \ref{fig:lines} illustrates a comparison between the number of code lines in our implementation and those in officially published versions. The majority of these paradigms can be efficiently implemented using Redco with only 100 to 200 lines of code. This efficiency boost of development can be attributed to Redco's ability to significantly reduce the need for writing boilerplate code.

\vspace{-5pt}
\section{Conclusion}
\vspace{-5pt}

In this work, we present a lightweight and user-friendly toolkit, \textit{RedCoast (Redco)}, designed to automate the distributed training of LLMs and simplify the ML pipeline development process. Redco incorporates an automatic model parallelism strategy, fundamentally based on two intuitive rules, without requiring additional coding efforts or MLSys expertise from the users. We evaluate its effectiveness on an array of LLMs, such as LLaMA-7B, T5-11B and OPT-66B. Furthermore, Redco has a novel and neat pipeline development mechanism. This mechanism requires users to specify only three intuitive pipeline design functions to implement a distributed ML pipeline. Remarkably, this mechanism is general enough to accommodate various ML algorithms and needs significantly fewer lines of code compared to their official implementations.

\newpage
\bibliography{custom}

\newpage
\appendix
\onecolumn

\section{Tensor Parallel Strategy Examples}

We provide examples of the generated sharding strategies by Redco toward different architectures, where \texttt{PartitionSpec('mp', None)} indicates partioning a parameter by dimension 0 and \texttt{PartitionSpec(None, 'mp')} indicates partioning a parameter by dimension 1, and \texttt{None} means saving a copy of the parameter across every device.

\begin{figure}[h]
    \centering
\begin{lstlisting}[language=Python]
params_sharding_rules_gptj = [
    (('fc_in', 'kernel'), PartitionSpec(None, 'mp')),  # Rule 1 in Section 3.1
    (('fc_out', 'kernel'), PartitionSpec('mp', None)),
    (('k_proj', 'kernel'), PartitionSpec(None, 'mp')), # Rule 2 in Section 3.1
    (('out_proj', 'kernel'), PartitionSpec('mp', None)), 
    (('q_proj', 'kernel'), PartitionSpec(None, 'mp')), 
    (('v_proj', 'kernel'), PartitionSpec(None, 'mp')), 
    (('(bias|scale)',), None),                         # Parameters other than transformer blocks or bias or scale terms
    (('embedding',), PartitionSpec('mp', None)), 
    (('lm_head', 'kernel'), PartitionSpec(None, 'mp')) 
]
\end{lstlisting}
    \vspace{-10pt}
    \caption{Sharding strategy for GPT-J generated by Redco.}
    \label{fig:enter-label}
\begin{lstlisting}[language=Python]
params_sharding_rules_llama = [
    (('up_proj', 'kernel'), PartitionSpec(None, 'mp')), # Rule 1 in Section 3.1
    (('gate_proj', 'kernel'), PartitionSpec(None, 'mp')), 
    (('down_proj', 'kernel'), PartitionSpec('mp', None)), 
    (('k_proj', 'kernel'), PartitionSpec(None, 'mp')),    # Rule 2 in Section 3.1
    (('o_proj', 'kernel'), PartitionSpec('mp', None)), 
    (('q_proj', 'kernel'), PartitionSpec(None, 'mp')), 
    (('v_proj', 'kernel'), PartitionSpec(None, 'mp')),
    (('(bias|scale)',), None),                            # Parameters other than transformer blocks or bias or scale terms
    (('embedding',), PartitionSpec('mp', None)), 
    (('lm_head', 'kernel'), PartitionSpec(None, 'mp')), 
    (('norm', 'weight'), None), 
    (('input_layernorm', 'weight'), None), 
    (('post_attention_layernorm', 'weight'), None)
]
\end{lstlisting}
\vspace{-10pt}
    \caption{Sharding strategy for LLaMA generated by Redco.}
    \label{fig:enter-label}
\end{figure}

\section{A Complete Distributed Training Example with Redco}

We provide a complete example for the distributed training of a T5-XXL model, which is able to run on multi-host environments. It uses a modeling from HuggingFace, on a summarization dataset, evaluated by rouge scores, and it saves the checkpoints with best rouge-2 and rouge-L scores. 

\begin{lstlisting}[label={lst:eval_summ}, language=Python]
from functools import partial
import fire
import numpy as np
import jax
import jax.numpy as jnp
import optax
import datasets
from transformers import AutoTokenizer, FlaxAutoModelForSeq2SeqLM
import evaluate
from redco import Deployer, Trainer


def collate_fn(examples,
               tokenizer,
               decoder_start_token_id,
               max_src_len,
               max_tgt_len,
               src_key='src',
               tgt_key='tgt'):
    model_inputs = tokenizer(
        [example[src_key] for example in examples],
        max_length=max_src_len,
        padding='max_length',
        truncation=True,
        return_tensors='np')

    decoder_inputs = tokenizer(
        [example[tgt_key] for example in examples],
        max_length=max_tgt_len,
        padding='max_length',
        truncation=True,
        return_tensors='np')

    if tokenizer.bos_token_id is not None:
        labels = np.zeros_like(decoder_inputs['input_ids'])
        labels[:, :-1] = decoder_inputs['input_ids'][:, 1:]
        decoder_input_ids = decoder_inputs['input_ids']
        decoder_input_ids[:, 0] = decoder_start_token_id
    else:
        labels = decoder_inputs['input_ids']
        decoder_input_ids = np.zeros_like(decoder_inputs['input_ids'])
        decoder_input_ids[:, 1:] = decoder_inputs['input_ids'][:, :-1]
        decoder_input_ids[:, 0] = decoder_start_token_id

    model_inputs['labels'] = labels
    decoder_inputs['input_ids'] = decoder_input_ids

    for key in decoder_inputs:
        model_inputs[f'decoder_{key}'] = np.array(decoder_inputs[key])

    return model_inputs


def loss_fn(train_rng, state, params, batch, is_training):
    labels = batch.pop("labels")
    label_weights = batch['decoder_attention_mask']

    logits = state.apply_fn(
        **batch, params=params, dropout_rng=train_rng, train=is_training)[0]

    loss = optax.softmax_cross_entropy_with_integer_labels(
        logits=logits, labels=labels)

    return jnp.sum(loss * label_weights) / jnp.sum(label_weights)


def pred_fn(pred_rng, params, batch, model, gen_kwargs):
    output_ids = model.generate(
        input_ids=batch['input_ids'],
        attention_mask=batch['attention_mask'],
        params=params,
        prng_key=pred_rng,
        **gen_kwargs)
    return output_ids.sequences


def output_fn(batch_preds, tokenizer):
    return tokenizer.batch_decode(batch_preds, skip_special_tokens=True)


def eval_rouge(examples, preds, tgt_key):
    rouge_scorer = evaluate.load('rouge')

    return rouge_scorer.compute(
        predictions=preds,
        references=[example[tgt_key] for example in examples],
        rouge_types=['rouge1', 'rouge2', 'rougeL'],
        use_stemmer=True)


def main(n_processes=None,
         host0_address=None,
         host0_port=None,
         process_id=None,
         n_local_devices=None,
         dataset_name='xsum',
         src_key='document',
         tgt_key='summary',
         model_name_or_path='facebook/bart-base',
         n_model_shards=1,
         n_epochs=2,
         per_device_batch_size=8,
         eval_per_device_batch_size=16,
         accumulate_grad_batches=2,
         max_src_len=512,
         max_tgt_len=64,
         num_beams=4,
         learning_rate=4e-5,
         warmup_rate=0.1,
         weight_decay=0.,
         jax_seed=42,
         workdir='./workdir',
         run_tensorboard=False):
    deployer = Deployer(
        n_model_shards=n_model_shards,
        jax_seed=jax_seed,
        workdir=workdir,
        run_tensorboard=run_tensorboard,
        n_processes=n_processes,
        host0_address=host0_address,
        host0_port=host0_port,
        process_id=process_id,
        n_local_devices=n_local_devices)

    dataset = datasets.load_dataset(dataset_name)
    dataset = {key: list(dataset[key]) for key in dataset.keys()}

    with jax.default_device(jax.devices('cpu')[0]):
        model = FlaxAutoModelForSeq2SeqLM.from_pretrained(
            model_name_or_path, from_pt=True)
        model.params = model.to_fp32(model.params)

        tokenizer = AutoTokenizer.from_pretrained(model_name_or_path)
        gen_kwargs = {'max_length': max_tgt_len, 'num_beams': num_beams}

    lr_schedule_fn = deployer.get_lr_schedule_fn(
        train_size=len(dataset['train']),
        per_device_batch_size=per_device_batch_size,
        n_epochs=n_epochs,
        learning_rate=learning_rate,
        schedule_type='linear',
        warmup_rate=warmup_rate)
    optimizer = optax.adamw(
        learning_rate=lr_schedule_fn, weight_decay=weight_decay)
    if accumulate_grad_batches > 1:
        optimizer = optax.MultiSteps(
            optimizer, every_k_schedule=accumulate_grad_batches)

    trainer = Trainer(
        deployer=deployer,
        collate_fn=partial(
            collate_fn,
            tokenizer=tokenizer,
            decoder_start_token_id=model.config.decoder_start_token_id,
            max_src_len=max_src_len,
            max_tgt_len=max_tgt_len,
            src_key=src_key,
            tgt_key=tgt_key),
        apply_fn=model,
        loss_fn=loss_fn,
        params=model.params,
        optimizer=optimizer,
        lr_schedule_fn=lr_schedule_fn,
        accumulate_grad_batches=accumulate_grad_batches,
        params_sharding_rules=deployer.get_sharding_rules(params=model.params))

    predictor = trainer.get_default_predictor(
        pred_fn=partial(pred_fn, model=model, gen_kwargs=gen_kwargs),
        output_fn=partial(output_fn, tokenizer=tokenizer))

    trainer.fit(
        train_examples=dataset['train'],
        per_device_batch_size=per_device_batch_size,
        n_epochs=n_epochs,
        eval_examples=dataset['validation'],
        eval_per_device_batch_size=eval_per_device_batch_size,
        eval_loss=True,
        eval_predictor=predictor,
        eval_metric_fn=partial(eval_rouge, tgt_key=tgt_key),
        save_last_ckpt=True,
        save_argmax_ckpt_by_metrics=['rougeL'])


if __name__ == '__main__':
    fire.Fire(main)
\end{lstlisting}

\end{document}